\documentclass[runningheads]{llncs}
\usepackage{amsmath}
\usepackage{booktabs} 
\usepackage{caption} 
\usepackage{subcaption} 
\usepackage{graphicx}
\usepackage{pgfplots}
\usepackage[all]{nowidow}
\usepackage[utf8]{inputenc}
\usepackage{tikz}
\usepackage{multicol}
\usepackage{algpseudocode,algorithm,algorithmicx}

\usepackage{graphicx}
\usepackage{hyperref}

\usepackage{mathtools}  
\usepackage{amsfonts}
\usepackage{amsfonts}
\usepackage{csquotes}

\pgfplotsset{compat=1.14}

\begin{document}
\title{Federated learning with incremental clustering for heterogeneous data}
%
%
\author{Fabiola ESPINOZA CASTELLON \and Aurélien MAYOUE \and
Jacques-Henri SUBLEMONTIER \and Cédric GOUY-PAILLER}

\authorrunning{F. ESPINOZA CASTELLON et al.}
%
\institute{Institut LIST, CEA, Université Paris-Saclay, F-91120, Palaiseau, France
\email{fabiola.espinozacastellon@cea.fr}}
\maketitle
\begin{abstract}
Federated learning enables different parties to collaboratively build a global model under the orchestration of a server while keeping the training data on clients' devices. However, performance is affected when clients have heterogeneous data. To cope with this problem, we assume that despite data heterogeneity, there are groups of clients who have similar data distributions that can be clustered. In previous approaches, in order to cluster clients the server requires clients to send their parameters simultaneously. However, this can be problematic in a context where there is a significant number of participants that may have limited availability. To prevent such a bottleneck, we propose FLIC (Federated Learning with Incremental Clustering), in which the server exploits the updates sent by clients during federated training instead of asking them to send their parameters simultaneously. Hence no additional communications between the server and the clients are necessary other than what classical federated learning requires. We empirically demonstrate for various non-IID cases that our approach successfully splits clients into groups following the same data distributions. We also identify the limitations of FLIC by studying its capability to partition clients at the early stages of the federated learning process efficiently. We further address attacks on models as a form of data heterogeneity and empirically show that FLIC is a robust defense against poisoning attacks even when the proportion of malicious clients is higher than 50\%.

\end{abstract}

\section{Introduction}
\label{sec:intro}
Federated Learning (FL) is a new distributed machine learning paradigm that enables multiple clients to build a common model under the orchestration of a central server. This paradigm functions while keeping the training data on clients' devices. McMahan et al. \cite{mcmahan2017communicationefficient}

introduced FL in order to preserve privacy and reduce the overhead communication costs due to data collection. Contrary to traditional server-side approaches which aggregate data on a central server for training, FL distributes learning tasks among clients and aggregates only locally-computed updates to build a single global model. Therefore, the global objective function $f$ to be minimized is formulated as a weighted sum of the local objective functions $f_k$:

\begin{equation}
  \min_{w}{f(w)}=\min_{w}{\sum_{k=1}^{K}{\frac{n_{k}}{\sum_{q=1}^K{n_q}}}f_k(w)}
\label{eq:weights}  
\end{equation}

\noindent
where each of the $K$ clients has $n_k$ samples and $\sum_{q=1}^K{n_q}$ is the total number of data points belonging to all clients. 

The local objective function $f_k$ measures the empirical risk over client-$k$'s local dataset. Its $n_k$ samples are drawn from a distribution $\mathcal{P}_k$:

\begin{equation}
f_k(w)=\mathbb{E}_{(x,y)\sim P_k}[l(w;x,y)]
\label{eq:loss}   
\end{equation}
where the local loss function $l(w;x,y)$ measures the error of the model $w$ in predicting a true label $y$ given an input $x$.

FL was formalized by the algorithm FedAvg \cite{mcmahan2017communicationefficient}. In FedAvg, the server randomly initializes a global model $w_0$, typically a deep neural network. At round $t$, the server selects a subset $C_t$ of $C \cdot K \leq K$ clients that take part in training and sends them the current global model $w_{t-1}$. Each participant $k$ runs several epochs of minibatch stochastic gradient descent to minimize its local loss function. Afterwards, each client sends back to the server its update $\delta_{t}^{k}$ that is the difference between $w_{t-1}$ and the optimized local parameters $w_{t}^{k}$. Finally, the server averages the received updates to obtain the global model $w_t=w_{t-1}-\sum_{k \in C_t}{\lambda_k \delta_{t}^{k}}$ where $\lambda_k = \frac{n_k}{\sum_{q \in C_t} n_q}$ is the weight associated to the client $k$,  thereby concluding a round of collaborative learning. The aggregation rule thus gives more weight in the weighted sum to clients having a higher number of examples. The FL process consists of multiple successive rounds.

\begin{figure}[t!]
  \centering
  \centerline{\includegraphics[width=8cm]{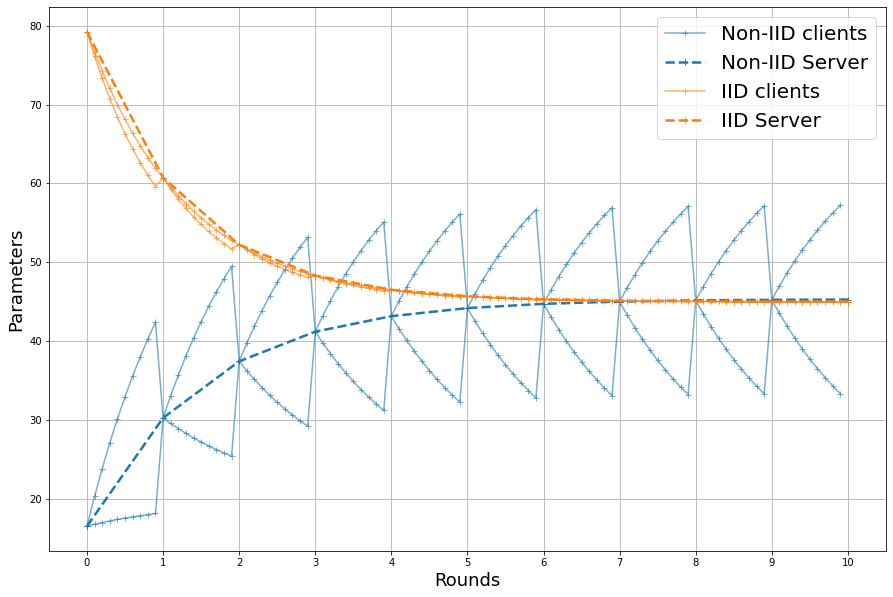}}
\caption{Linear regression example: evolution of clients' weights during federated training for IID and non-IID cases. We notice that non-IID clients' weights diverge, whereas IID clients converge to the same optimal weight.}
\label{fig:divergence}
\end{figure}

Throughout the learning processes, the independent and identically distributed (IID) sampling of training data is a key point for training accurate models. It ensures that the stochastic gradient is an unbiased estimate of the full gradient. However, in FL scenarios where clients generate personal data from different locations and environments, it is unrealistic to assume that clients' local data is IID, i.e., each client's local data is uniformly sampled from the entire training dataset composed of the union of all local datasets. In non-IID scenarios, the global performance of FedAvg is severely degraded \cite{zhao2018federated} because the heterogeneity of data distribution across clients results in weight divergence during the collaborative training. At a round $t$, the difference between the data distribution of two clients $i$ and $j$ causes locally trained weights $w_t^i$ and $w_t^j$ to diverge and the convergence rate, precision and fairness of the federated model to degrade by comparison with homogeneous data. 
Figure \ref{fig:divergence} illustrates this phenomenon, simulating here a simple case of $1$-D linear regression in a FL context with two clients. In this toy problem, each client performs $10$ local epochs with $50$ samples, and the server executes $10$ global rounds. In the IID case, the clients collaborate to infer the same parameter equal to $45$. In the non-IID case, the parameters to be inferred of the first and second client are $20$ and $70$ respectively. For the IID case, both clients' weights follow the same direction and converge to the \textit{same} optimum, whereas for the non-IID case, clients' weights point to different directions and make the global model converge to a parameter different from their own optimums, which is the center of both parameters.

When non-IID is mentioned in the FL setting, it typically means that for two clients $i$ and $j$, $\mathcal{P}_i \neq \mathcal{P}_j$. Based on \cite{hsieh2020noniid,kairouz2021advances}, and knowing that for client $i$, $\mathcal{P}_i(x,y) = \mathcal{P}_i(y| x)\mathcal{P}_i(x) = \mathcal{P}_i(x| y) \mathcal{P}_i(y)$, different cases of non-IID data can be distinguished. Concept shift cases occur when conditional distributions vary across clients: 
\begin{itemize}
    \item \textit{Concept shift on features:} marginal label distributions are shared $\mathcal{P}_i(y) = \mathcal{P}_j(y)$ but conditional distributions vary across clients $\mathcal{P}_i(x|y) \neq \mathcal{P}_j(x|y)$. This can arise in handwriting because some people might write \textquote{7} with bars or without and so, features might be different for a same label (number). We can also refer to this case as \textquote{different features, same labels}.
    \item \textit{Concept shift on labels:} marginal features distributions are shared $\mathcal{P}_i(x) = \mathcal{P}_j(x)$ but label distributions conditioned on features vary across clients $\mathcal{P}_i(y|x) \neq \mathcal{P}_j(y|x)$. This can occur in sentiment analysis : for the same features, people can have different preferences (labels). This case can be referred as \textquote{same features, different labels}. It is also illustrated by the non-IID case in Figure~\ref{fig:divergence} because for the same inputs i.e. features, linear models will have different results i.e labels because their parameters are different (in this example the parameters are $20$ and $70$). 
\end{itemize}

This paper focuses on concept shift cases that can be addressed by clustering, deliberately omitting cases where marginal distributions vary across clients, because, on the one hand, machine learning is inherently robust to feature distribution skew ($\mathcal{P}_i(x) \neq \mathcal{P}_j(x)$ when $\mathcal{P}(y|x)$ is shared). Typically, one of the advantages of a convolutional neural network is to be robust to variant features through convolutions and pooling. On the other hand, clustering data with label distribution skew ($\mathcal{P}_i(y) \neq \mathcal{P}_j(y)$ when $\mathcal{P}(x|y)$ is shared) would group clients who only have a certain number of labels. Methods inspired by the incremental learning literature (FedProx \cite{Li2018}, SCAFFOLD \cite{Karimireddy2020} and SCAFFNEW \cite{mishchenko22}) are more suitable to address this latter case. 

Another setting that can degrade the performance of a federated model is when it is attacked by malicious clients that try to poison the model during training-time \cite{mhamdi2018hidden}. Under the strong assumption that a malicious client $k$ has full knowledge of the aggregation rule used by the server and of the updates of others, it can make the aggregation result equal to an arbitrary value $U$ at any round $t$ by submitting the following update:
\begin{equation}
\delta_k^t = \frac{1}{\lambda_k} U  - \sum_{i \in C_t, i\ne k}\frac{\lambda_i}{\lambda_k} \delta_i^t
\label{eq:poison}    
\end{equation}

FedAvg, and more specifically the mean aggregation rule, are inherently vulnerable to these attacks, as shown by (\ref{eq:poison}). However, in practice clients do not have a full knowledge of the system. That is why the standard model poisoning attacks often consist in sending an update containing random weights, null weights, or, more efficiently, the opposite of the true weights. The impact of the attack can also be strengthened when several clients collude with each other. Such attacks can be considered as a form of data heterogeneity because the poisoned updates are different of other updates as they try to hinder the global model convergence.

\section{Related work}\label{sec:work}
Improving FL models while dealing with non-IID data is an active field of research. Most of the approaches in literature try to personalize the global FL model to improve performances of individual clients. In works based on transfer learning \cite{chen2021fedhealth,yu2020salvaging} and meta-learning \cite{jiang2019improving}, the global model is trained using FedAvg and afterwards each client fine-tunes the shared model using its local data. In multi-task learning, the clients' models are trained simultaneously by exploiting commonalities and differences across the learning tasks. MOCHA \cite{smith2018federated} uses the correlation matrix among tasks as a regularization term while FedEM \cite{marfoq21} considers that the data distribution of each client is a mixture of unknown but shared underlying distributions and uses the Expectation-Maximization algorithm for training.

\begin{figure}
  \centering
  \centerline{\includegraphics[width=17cm,height=5cm,keepaspectratio]{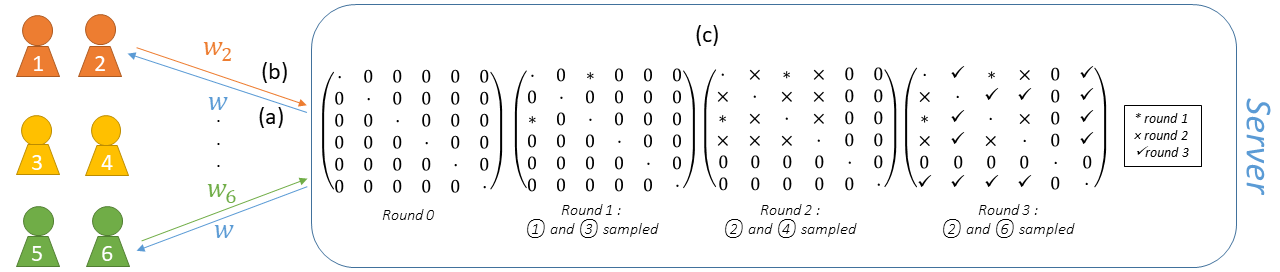}}
\caption{An overview of Incremental Clustering. (a) The server initializes the model. (b) Clients perform local training and send their parameters to the server who keeps them in memory. (c) It then computes the similarity between each parameter it has access to and fills in the adjacency matrix. The parameters received at that round are then averaged and sent back to clients. For example, at round 1, client 1 and 3 are sampled, so we can compute $S_1^{1,3}=S_1^{3,1}$. At round 2, client 2 and 4 are sampled, which means we can compute $S_2^{2,4}=S_2^{4,2}$ but also $S_2^{1,2}, S_2^{2,3}, S_2^{1,4}$ and $S_2^{3,4}$. At round 3, since client 2 was already sampled, we replace its parameters kept in memory by the most recent. The similarities linked to client 2 are thus recomputed, and the ones of client 6 are computed as well.}
\label{fig:ic}
\end{figure}

FedAvg permits to collaboratively learn a unique model while personalized approaches provide one model per client. We consider that clustering-based methods can be a relevant compromise between collaboration and personalization. Several works \cite{Ghosh,sattler2019clustered,Briggs} have already considered that it is possible to find a cluster structure in order to gather clients with similar data distributions and perform classical FedAvg training per cluster. Thus, there are as many models as clusters. 

In \cite{Ghosh}, the number of clusters are estimated \textit{a priori} and each client is assigned to one of them before performing local training. Once the cluster of each client is identified, the server averages the parameters of each clusters separately. Determining each client's cluster requires high communication costs and may be unsuitable for large deep learning models.

Our work resembles more the one of \cite{sattler2019clustered,Briggs} who cluster clients based on their model parameters after FedAvg training. 
However, these approaches perform a communication round $T$ involving \textit{all} clients to build the clusters. In a cross-device setting where the number of clients is considerable \cite{hard2019federated}, this step is impractical in terms of clients’ availability and communication costs. To prevent such a bottleneck and to adapt to real-world applications, our method takes advantage of the updates received during FedAvg rounds and builds an adjacency matrix incrementally as clients are sampled for training.

Concerning model attacks, existing methods try to prevent the influence of the malicious clients by replacing the averaging step on the server-side with robust estimates of the mean, such as coordinate-wise median \cite{yin2021byzantinerobust} or Krum, an aggregation rule based on a score using couples of closest vectors  \cite{blanchard2017machine}. However, theses approaches remain robust to model poisoning attacks while the proportion of adversaries that participate in each round of learning is strictly below 50$\%$ \cite{hu2021challenges}. In Section \ref{sec:Exp}, we will compare our cluster-based approach to the median aggregation rule method \cite{yin2021byzantinerobust}, an efficient defense which we will refer to as \textit{median defense}.

\section{Incremental clustering} \label{algo}

\begin{figure}[t!]
  \centering
  \centerline{\includegraphics[width=8cm]{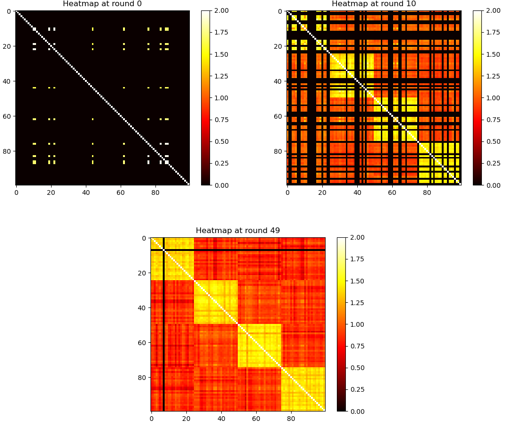}}
\caption{Evolution of the similarity matrix through rounds for an example with four clusters and for parameters $K=100$ and $C=0.1$ throughout rounds $0$, $10$ and $49$. At round $0$ the matrix contains $10 \times 10$ non-zero values because only $10$ clients have been sampled. As the server samples more clients, the matrix fills up. Calculated similarities change during training if clients are re-sampled. At round $49$, one of the clients has never been sampled. Thus, the graph resulting from the similarity matrix of round $49$ will not contain that client in its nodes, and so the client will not be assigned to a cluster.}
\label{fig:heatmaps}
\end{figure}

Similarly to prior works, we tackle the issue of heterogeneous data by extending FedAvg and adding a clustering step to separate clients into groups. We next train them independently to reach homogeneous data performance. However, contrary to \cite{sattler2019clustered,Briggs}, we avoid performing the burdensome round mentioned in Section \ref{sec:work} by taking advantage of the local updates we already have access to at each round. Specifically, at round $t$ of FedAvg, when $|C_t|$ clients finish local training, each client $k$ sends to the server its updates $\delta_t^k=w_{t-1}-w_t^k$. This is a good indicator of how clients' weights differ from the global model. As seen in Algorithm \ref{alg:IC} (l.\ref{line:s}) and Figure \ref{fig:ic}, these values are stored by the server in a matrix $M$ in order to fill in an adjacency matrix $S_t$ afterwards. This matrix contains the similarities between clients : $S_t=\left(s(\delta^i, \delta^j)\right)_{1 \leq i,j \leq K}$ where $s(\delta^i, \delta^j)$ is the similarity measure between the updates of client $i$ and $j$.
During next round $t+1$, the server stores the new updates of clients belonging to $C_{t+1}$. In order to compute the most similarities between clients, the server calculates $s$ between recent and previous updates kept in memory. To this end, it keeps the most recent update if a client has already been sampled and does not forget previously stored updates. If a client has never been sampled, the values of its corresponding row and column in $S_t$ will be equal to zero. Furthermore, if a similarity between two clients has already been computed, we keep the most recent one (see Figure \ref{fig:heatmaps}). 

Once FedAvg stabilizes, the second part of Algorithm \ref{alg:IC} (l.\ref{line:louvain}) begins : we cluster clients by creating a graph from $S_t$ and applying the Louvain method algorithm \cite{Blondel_2008}. Once we have detected different communities, that we call clusters, we resume separately FedAvg per clusters. Within a cluster, we expect that clients should have the same data distribution and the performances should reach the ones of identically distributed data.

It should be noted that our incremental method adds two biases in comparison with approaches requiring all clients to take part in the same round:
\begin{itemize}
    \item The coefficients of $S_t$ are not all calculated at the same round because the matrix fills in during rounds. Thus, at round $t$, the computed similarities are added to $S_t$ which also contains similarities computed at previous rounds $\tau, \tau<t$.
    
    \item $s$ is often computed for updates of different rounds. For instance, if client $i$ was sampled at round $t$ and client $j$ at round $\tau<t$, then $S_{t}^{i,j}$ will be equal to $s(\delta_{t}^i, \delta_\tau^j)$. 
\end{itemize} 

In previous methods \cite{sattler2019clustered,Briggs}, similarities are computed for a same round. Typically, the similarity between client $i$ and $j$ would be $s(\delta_{t}^i, \delta_t^j)$. We notice that the gap between our approach and the benchmark lies on the difference between the updates of client $j$ computed at different times.
To clarify this, let us consider for simplicity that clients perform a single local epoch $E$ and SGD with a learning rate $\alpha$. Common similarities (Euclidean, Manhattan, Minkowski,...) between two points are associated with distances defined as the $p$-norm of the difference between these points for a certain $p$. 

Previous methods \cite{sattler2019clustered,Briggs} would compute distances between updates of two distinct clients calculated at the same round $t$ as follows: 
\begin{equation}
     \|\delta_t^i -\delta_t^j\|_p =\|(w_{t-1}-w_t^i)-(w_{t-1}-w_t^j)\|_p =\|w_t^j-w_t^i\|_p \label{eq:them}\\
\end{equation}

However, our method computes distances of updates obtained at different rounds, for instance $t$ and $\tau<t$. Thus:
\begin{equation}
\begin{split}
    \|\delta_{t}^i - \delta_\tau^j \|_p &= \|(w_{t-1}-w_t^i)-(w_{\tau-1}-w_\tau^j)\|_p \\
    &= \|(w_{t-1}-w_{\tau-1})+(w_\tau^j-w_t^i)\|_p \label{eq:ours}\\
    &= \|(w_{t-1}-w_{\tau-1})+(w_t^j-w_t^i)+(w_\tau^j-w_t^j)\|_p \\
\end{split}
\end{equation}

To upper bound the difference between the norms of the benchmark (\ref{eq:them}) and ours (\ref{eq:ours}), we use the triangle inequality : 
\begin{equation}
\begin{split}
\|\delta_{t}^i - \delta_\tau^j \|_p - \|\delta_t^i -\delta_t^j\|_p &\leq \|(w_{t-1}-w_{\tau-1})+(w_\tau^j-w_t^j)\|_p \\
     &\leq \underbrace{\|(w_{t-1}-w_{\tau-1})\|_p}_{\text{(a)}} + \underbrace{\|(w_\tau^j-w_t^j)\|_p}_{\text{(b)}} \label{eq:diff}\\
\end{split}
\end{equation}

As we will discuss in Section \ref{sec:Exp}, the difference between our method and previous ones mainly relies on the sampling of clients during the training process. To get a better sense of term (\ref{eq:diff}.a), we can notice by induction that :
$$w_{t-1}=w_{t-2} - \sum_k \lambda_k \delta_{t-1}^k =...=w_{\tau-1} - \sum_{t'=\tau}^{t-1} \sum_{k \in C_t'} \frac{n_k}{\sum_{q \in C_t'} n_q} \delta_{t'}^k$$

Thus, if client $j$ was sampled at a round much earlier than client $i$, term (\ref{eq:diff}.a) can be large for two reasons. Firstly, within a same round $t'$ the heterogeneity of data causes divergence, as we mentioned in section \ref{sec:intro}. Moreover, the more rounds take place in between time $t$ and $\tau$, the more terms are added. We can note in Figure \ref{fig:divergence} that the difference between global weights becomes more significant if the rounds are distant. However, if FedAvg reaches convergence, $w_{t-1}$ and $w_{\tau-1}$ will likely be similar, making term (\ref{eq:diff}.a) negligible. We can again notice in Figure \ref{fig:divergence} that at a stabilized state (presumably after round $8$ for this toy example), global weights will be comparable and thus their difference small.

Term (\ref{eq:diff}.b) also depends on when client $j$ was sampled. Although training is done on the same local data for client $j$, $w_\tau^j$ and $w_t^j$ can be significantly different if their starting points are distant.

Despite these biases, we use the updates received at each round to perform clustering because we consider that they contain relevant information about the clients' direction in the optimization process. 

Given that we consider a cross-device context with a high number of clients, it is possible that during federated training not all clients will be sampled. If it is the case, no value will be present in its corresponding row and column in $S_t$, thus it will not be clustered. Following \cite{Ghosh}, to assign them to a group, we evaluate each model of the clusters with their test data. We notice that the highest accuracy is obtained by the model corresponding to the clients' clusters.

Note that $S_t$ does not influence in the computing of FedAvg and that we perform no additional communications between the server and the clients than what classical FL requires. The central server performs the supplementary calculations due to the adjacency matrix. We consider that in a decentralized setting where the server computational capacity is significantly higher than the clients', this extra work is not critical. 

As we will see in Section \ref{sec:Exp}, our method can tackle the statistical challenge inherent to FL and find the correct clustering structure for different cases of non-IID data. It also avoids the step in which all clients send their updates to the server, which is very demanding in terms of communications costs. 

Furthermore, by addressing the security challenge, our objective is to separate the updates from malicious and loyal clients and then, to build a global model from a cluster containing only loyal clients. We show in Section \ref{sec:Exp} that our approach is a robust defense even if there are a majority of adversaries.
\begin{algorithm}[!t]
\caption{FL through incremental clustering. $T$ rounds of FedAvg are performed before clustering and $T_f$ rounds after. $S_t$ is the adjacency matrix at round $t$ and matrix $M$ stores clients updates.}\label{alg:IC}
\begin{algorithmic}[1]
\Procedure{FLIncrementalClustering}{$K$}
\State initialize $w_0$
\For{$t$ in $t=1,...,T$}
\State $C_t \gets$ random subset of all clients $K$

\For{each client $k$ in $C_t$}
\State $\delta^k_{t}, n_k$ = \small \textsc{ClientUpdate($w_{t-1}, k, E, B, \alpha$)}
\State $M_k = \delta^k_{t}$ \Comment{Server stores update} \label{line:s}
\EndFor

\State $w_{t}= w_{t-1} - \sum_k \frac{n_k}{\sum_{q \in C_t}{n_q}}\delta^k_{t}$

\For{$i, j$ in $K$} 
\State $S_t^{i,j}=s(M_i, M_j)$ \Comment{Update $S_t$ matrix}
\EndFor
\EndFor

\State $P \gets \small \textsc{LouvainMethod}(S_T)$ \label{line:louvain}

\For{cluster $c$ in $P$} \label{line:louvain2}
\State initialize server $c$ with weights $w_{c,T}=w_T$

\For{$t$ in $t=T+1,..., T+T_f$}
\State $C_{c,t} \gets$ random subset of clients in cluster $c$

\For{each client $k$ in $C_{c,t}$}
\State $\delta^k_{t}, n_k$ = \small \textsc{ClientUpdate}($w_{c,t-1}, k, E, B, \alpha$)
\EndFor

\State $w_{c,t}= w_{c,t-1} - \sum_k \frac{n_k}{\sum_{q \in C_t}{n_q}}\delta^k_{t}$
\EndFor
\EndFor \label{line:louvain3}
\EndProcedure

\Procedure{ClientUpdate}{$w, k, E, B, \alpha$}
\State initialize $w_k= w$
\For{$e$ in $e=1,..., E$}
\State Divide $n_k$ samples into batches of size $B$ : set $\mathcal{B}$

\For{b in $\mathcal{B}$}
\State $w_k = w_k - \alpha \nabla l(w_k, b)$
\EndFor
\EndFor
\State $\delta_k = w-w_k$
\State \Return $(\delta_k, n_k)$ \Comment{\small Send update and number of samples}
\EndProcedure
\end{algorithmic}
\end{algorithm}
\section{Experiments ans discussion} \label{sec:Exp}

\subsection{Dataset and model} \label{Datasets}

We propose to use the community detection algorithm Louvain method \cite{Blondel_2008} for clustering. Contrary to \cite{Ghosh,sattler2019clustered,Briggs}, we want to avoid having a cluster dependent parameter because we assume that we do not know \textit{a priori} the cluster structure or the number of clusters we can expect. Louvain is a greedy algorithm that does not require such parameter. 
The basic Louvain algorithm considers graphs with positive weights. To this end, we define the following similarity, positive, upper bounded by $2$ and that uses the cosine distance as in \cite{sattler2019clustered}, which gives a good sense of the \textit{direction} gradients take during training:
\begin{equation}
\label{eq:sim}
s(\delta^i,\delta^j)=1+cos(\delta^i,\delta^j) 
\end{equation}
Our approach is however agnostic to the clustering method and could be applied with other clustering algorithms.

We realize two types of experiments. The first ones simulate non-IID cases by artificially forming groups of clients with similar properties. These cases are \textit{verifiable} in the sense that after we apply our method, we can verify if clients following same data distributions are effectively grouped together. We will refer to this as the \textit{correct} clustering. 
Experiments are done with the MNIST dataset \cite{lecun2010mnist} dedicated to identify handwritten digits from pixel data. The dataset is partitioned into 100 clients, each having 600 training samples and 100 test samples. We simulate a user experience context with heterogeneous data by forming groups of clients following different data distributions. The non-IID case \textquote{same label, different features} is simulated by partitioning the data into four groups. Within each group the images are rotated of 90 degrees.

We will refer to this experiment as \textit{image rotation}. Similarly, for the non-IID case \textquote{same features, different labels}, we partition the clients into 5 groups and within each group two digit labels are swapped. This will be referred to as \textit{label swap}. For instance, a client of the first group will have 0 and 1 images labeled with 1 and 0 respectively. A client of the second group will have the correct labels for 0 and 1 images but will have labels 3 and 2 for 2 and 3 images.

The second type of experiments concern the application of our method as a defense to poisoned models. We continue to use the MNIST dataset \cite{lecun2010mnist} split into $100$ clients, from which a certain number of them will be attackers. An attacker is a malicious client that sends $-\delta_t^k$, \textit{i.e.} the opposite of its true update, in order to cause the global model to diverge. This attack will be referred to as \textit{minus grad attack}. The more clients are attackers, the more the global model will be perturbed. We set the strength of the attack by varying the number of adversaries and we realize experiments for $30$, $40$, $50$ and $60$ malicious clients out of the $100$ total clients. We implement an existing defense that replaces the typical mean aggregation by a median aggregation \cite{yin2021byzantinerobust} in order to compare its results to the ones of FLIC as a defense. As in the previous experiments, we can also speak of a \textit{correct} clustering if all malicious clients are separated from loyal clients. It should be noted that our method is agnostic to the aggregation rule. In our experiments we use the weighted mean, but a median aggregation could be used in order to combine both defenses.

Following \cite{mcmahan2017communicationefficient}, we build a convolutional neural network of two convolutional layers followed by a fully connected layer with a ReLu activation and a final softmax output layer. This architecture is used by both the server and the clients. We perform SGD with a learning rate of $\alpha=0.01$. Unless specified otherwise, the number of local epochs $E$ and the size of mini-batches $B$ are 5 and 10 respectively.

We used GPUs with specifications INTEL Skylake AVX512 support, 192Go RAM and at least 48 threads. The wall-clock time of computation for parameters $E=5$, $B=10$, $C=0.1$ and $200$ rounds was in average of one hour.

\subsection{Results on the statistical challenge}
In this section, we lead two sets of experiments realized 20 times each to reduce randomness. During our first set of experiments, we assess the performance of our method when clients have heterogeneous data. Firstly, 10\% of the clients are randomly sampled at each round. Then, at round 200 when FedAvg reaches convergence, we cluster clients thanks to the adjacency matrix built during the 200 rounds.

After clustering, each group performs 5 rounds of FedAvg. 

\begin{figure}[t!]
\begin{minipage}[b]{1.\linewidth}
  \centering
  \centerline{\includegraphics[width=9.5cm]{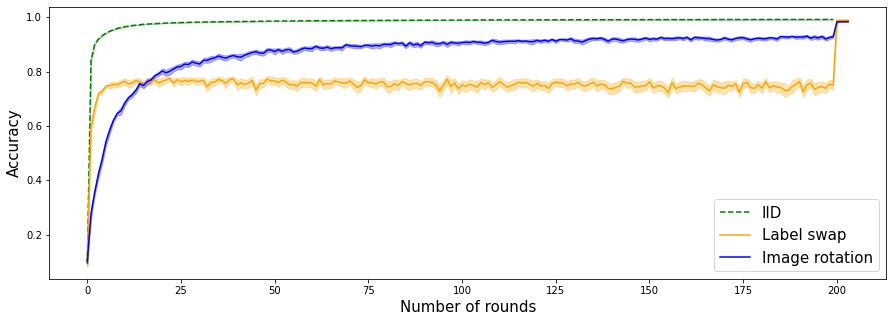}}
  \centerline{(a) Accuracy before clustering}\medskip
\end{minipage}
\begin{minipage}[b]{.48\linewidth}
  \centering
  \centerline{\includegraphics[width=4.5cm]{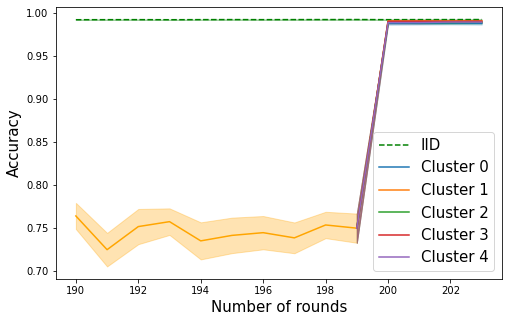}}
  \centerline{(b) Label swap}\medskip
\end{minipage}
\hfill
\begin{minipage}[b]{0.48\linewidth}
  \centering
  \centerline{\includegraphics[width=4.5cm]{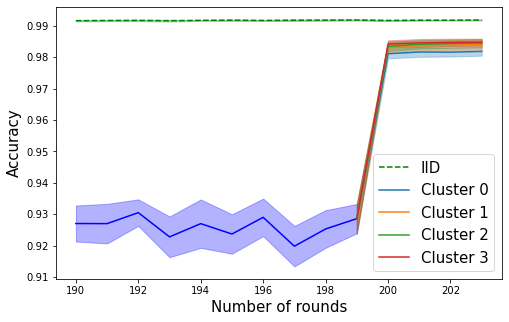}}
  \centerline{(c) Image rotation}\medskip
\end{minipage}
\caption{(a) FedAvg accuracy for IID, label swap and image rotation cases before and after clustering. (b)-(c) Focus on cluster's accuracies trained for 5 rounds for label swap and image rotation cases respectively. The displayed results are obtained by evaluating the servers' models on test images of clients managed by them. We display the mean result of all experiments with $0.95$ confidence interval.}
\label{fig:res1}
\end{figure}

\begin{table}[t!]
\centering
\begin{tabular}{c|c|c}
\hline
\multicolumn{1}{c|}{}& Pre-clustering & Post-clustering \\
\hline
IID data & \multicolumn{2}{c}{$0.99\pm 0.01$} \\
\hline
Label swap & $0.75\pm 0.075$ & $0.99 (\times1.32)\pm 0.011$ \\ 
\hline
Image rotation & $0.93\pm 0.051$ & $0.98(\times1.05)\pm 0.013$ \\
\hline
\end{tabular}
\caption{Accuracy evolution before and after clustering. We display the mean of clients' accuracies $\pm$ the standard deviation and in parenthesis the relative increase from before clustering.}
\label{tab:prepost}
\end{table}

For both non-IID cases, all clients were correctly clustered as in \cite{Briggs} but without the necessity of sampling all clients at round $200$.
Figure \ref{fig:res1} shows that after clustering, the mean accuracy of the models increases of $24\%$ and $5\%$ for the label swap case and the image rotation case respectively. We reach the same performances as in the IID case, which proves our method is well adapted to tackle the problem of heterogeneous data. Finally, Table \ref{tab:prepost} indicates that the variability between clients' results has decreased because the standard deviation after clustering decreases. This contributes to improve the fairness of the federated models.

\begin{figure}[t!]
  \centering
  \centerline{\includegraphics[width=7cm]{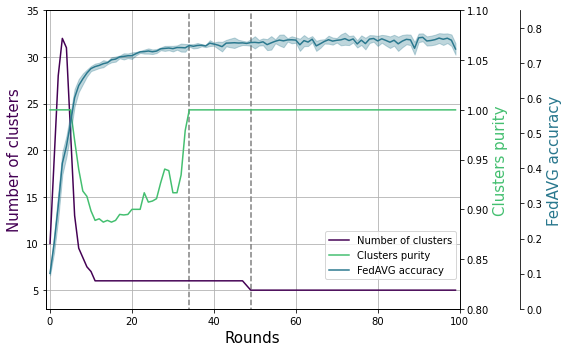}}
\caption{Mean FedAvg accuracy with $0.95$ confidence interval for parameters $E=3$ and $B=50$ over $100$ rounds (blue curve and outer right axis). Number of clusters (violet curve and left axis) and clusters purity if (green curve and inner right axis) if clustering is done with the similarity matrix of each round of the abscissa.}
\label{fig:conv}
\end{figure}

For our second set of experiments (Figure \ref{fig:conv}), we evaluate the capability of our method to well cluster heterogeneous data at early rounds i.e. before convergence. 

To this end, we cluster clients at every round - without training per cluster as in Algorithm \ref{alg:IC} (l.\ref{line:louvain2}-\ref{line:louvain3}).
We focus only on the label swap case and change parameters $E$ and $B$ to $3$ and $50$ in order to slow down the convergence rate and to better analyze the behavior of Algorithm \ref{alg:IC}.

In Figure \ref{fig:conv}, we plot in green the cluster purity. We define this quantity as the percentage of clients grouped with others who follow the same data distributions. If it is equal to $1$, it means that FLIC formed groups of clients who all follow the same data distribution. 

We notice in Figure \ref{fig:conv} that at the beginning of the training, the cluster purity is equal to $1$ and the number of clusters represented by the violet curve starts at $10$.

At the beginning of the training, the graph is small because not many clients have yet been sampled. For instance at round $1$, since $C=0.1$, the graph contains only $10$ nodes with similar edges. The Louvain method can not seam to find clear communities among nodes and forms $10$ communities each containing a single client, which explains the obtained purity. 

As new clients are sampled and trained, more communities are found, but they sometimes contain clients of different data distributions, which makes the cluster purity decrease from its optimal value $1$. Before round $34$, the lowest round for which a client $k$ was sampled is in average equal to $1$ i.e. $\delta_1^k$ is used for the update of the adjacency matrix. Some clients have thus performed few epochs of local training and their updates are not clearly distinguishable. After round $34$, clusters purity reaches $1$, so performing training per cluster can enhance performances as clients per group follow same data distributions. Yet, these partitions are not optimal as not \textit{all} clients with same distributions are grouped together. As shown in Figure \ref{fig:conv} by the violet curve, starting from round $49$, five pure groups are found, making the clustering correct. At this point, the mean lowest round at which a client was sampled is equal to $6$. Most clients have thus updates that are representative of their local objective functions. 

We can thus leverage the information received by the clients' updates during federated training even if they are computed at an early stage of their local optimization. 

In practice, the updates sent by clients at early rounds could be ignored for the clustering.

\begin{figure}[t!]
  \centering
  \centerline{\includegraphics[width=9.3cm,keepaspectratio]{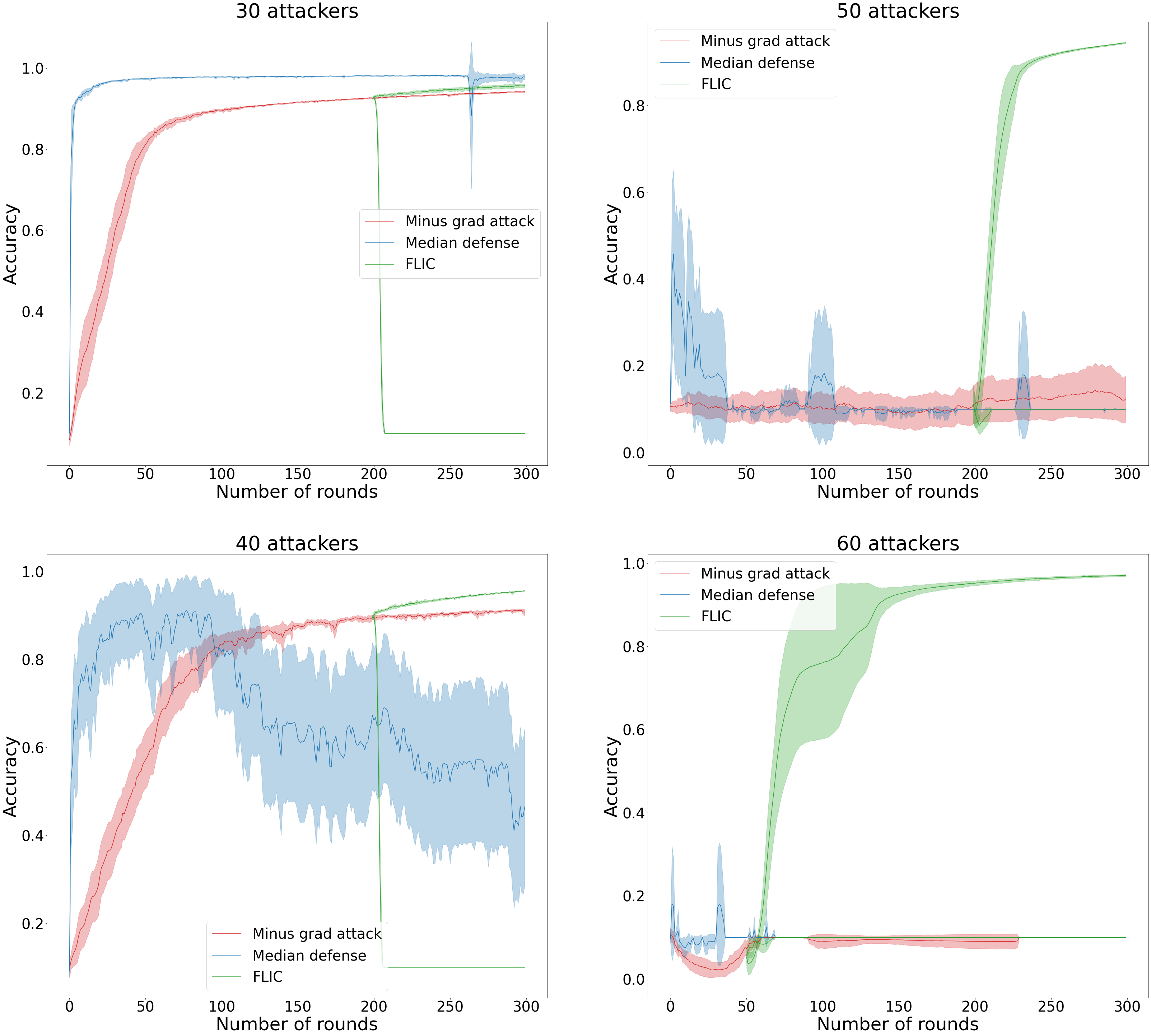}}
\caption{Minus grad attack by $30$, $40$, $50$ and $60$ attackers out of $100$ loyal clients, defended by the median defense and FLIC. The displayed results are obtained by evaluating the servers’ models on test images of the clients managed by them. For the FLIC defense, clusters containing malicious clients are evaluated on clean images, which explains the drop in accuracy for them. We display the mean result of all experiments with $0.95$ confidence interval.}
\label{fig:attacks}
\end{figure}
\subsection{Results on the security challenge}

\begin{table}[!t]
\begin{center}
\resizebox{\columnwidth}{!}{%
\begin{tabular}{ c || c| c| c| c}
  \hline
  & 30 attackers & 40 attackers & 50 attackers & 60 attackers\\ 
  \hline
  \hline
  FedAvg without atack & \multicolumn{4}{c}{$0.99\pm 0.01$}\\ 
  \hline
  \hline
  FedAVG with attack & $0.94\pm0.003$ & $0.91\pm0.005$ & $0.14\pm0.14$ & $0.1\pm0.00$ \\
  \hline
  \hline
  Median defense & $0.98\pm0.002$ & $0.57\pm0.423$ & $0.10\pm0.004$ &$0.1\pm0.00$ \\
  \hline
  \hline
  FLIC - loyal clusters & $0.95\pm0.012$ & $0.94\pm0.004$ & $0.91\pm0.006$ & $0.97\pm0.004$\\
\hline
Mean number of loyal clusters & $3.5$ & $1.09$ & $1$ & $1.1$\\
\hline
Mean number of clients in loyal clusters & $20$ & $55$ & $50$ & $40$\\
\hline
\end{tabular}%
}
\end{center}
\caption{Effect of minus grad attack on IID data and performances of both median and FLIC defenses according to the number of total malicious clients. For every method we display the best accuracy of the last 50 rounds $\pm$ the corresponding standard deviation.  Results on FLIC defense concern only clusters of loyal clients. The displayed results are the accuracy after clustering, the mean number of loyal clients and the mean number of clients inside a loyal cluster.}
\label{tab:att_pur}
\end{table}

We also realize $20$ experiments and randomly sample $10\%$ of the clients at each round,  with local parameters $E=1$ and $B=50$. We implement the attack mentioned in Subsection \ref{sec:Exp}.A, which is referred to as \textit{minus grad attack}. We evaluate our method FLIC as a defense and compare our results with the coordinate-wise median aggregation rule defense \cite{yin2021byzantinerobust}.

During FLIC training, we expect that after clustering, malicious clients are separated from loyal. The results in Figure \ref{fig:attacks} and Table \ref{tab:att_pur} are evaluated on client's test data. A malicious client will thus have poor performances if it is evaluated on its own corrupted model. If FLIC is a robust defense, its plot should split into two curves at round $T$, one representing loyal clients reaching adequate performances, and the other representing malicious clients dropping to a poor accuracy. 

All simulations perform $300$ total rounds for the minus grad attack and the median defense. For the experiments on FLIC, the first simulations with $30$, $40$ and $50$ attackers, run $T=200$ rounds before clustering and $T_f=100$ rounds after clustering. We decided to perform more rounds after clustering than in the previous experiments in order to see if the new model, cleaned of malicious clients, can reach performances of models who have not been attacked. 

As we can see in Figure \ref{fig:attacks} and Table \ref{tab:att_pur}, when the number of malicious clients is not high, for instance $30$, the median defense is robust and slightly outperforms FLIC. However, when the number of attackers is equal to $40$, the median defense learns with difficulty the task at the beginning of the training, and then fails after approximately $100$ rounds. As mentioned, the median defense is robust only if the majority of clients are loyal \cite{hu2021challenges}. At a particular round, since there are $40\%$ of attackers, it is possible that out of the $10$ sampled clients, more than $5$ will be attackers, which makes the median defense fail. The variability of the median defense for $40$ attackers in Figure \ref{fig:attacks} is thus due to the changing number of attackers at each round and for every simulation. On the contrary, as FLIC correctly separates clients at round $T=200$, loyal clients reach good performances again.

When the proportion of attackers is of $50\%$ or higher, the effects of the minus grad attack become predominant and the median defense collapses. FLIC manages though to correctly separate clients. For $50$ attackers, the new model after clustering reaches good performances in less than $100$ rounds. 

For the experiment with $60$ attackers, we performed less rounds before clustering ($T=50$) because we noticed that if done later (for instance at round $T=200$ as the other experiments), when FLIC successfully separated malicious clients of the rest, loyal clients restarted training with a model that was too degraded by the attack and could not learn the objective task in the remaining rounds. If clustering is performed at round $T=50$, the new model of loyal clients has a lower convergence rate but the starting point of loyal clients' training is not as distant from their objective as before ($T=200$) and they can still rebuild an efficient model.

Clusters purity defined in Subsection \ref{sec:Exp}.B now reflects the capability of FLIC to correctly split malicious and loyal clients. If a malicious client is grouped with loyal clients, it decreases the methods purity. For all of our experiments, FLIC clusters correctly clients (purity equal to $1$).

In Table \ref{tab:att_pur} we display the mean number of loyal clusters and the mean number of clients in loyal clusters. Ideally, \textit{all} loyal clients should be grouped in one single cluster, in order to enhance the collaborative characteristic of FL. With $30$ attackers, loyal clients are generally not grouped all together but rather in small clusters of $20$ clients. Under this attack, the model can still learn the training task, and thus clients can more easily learn their own local objective before sending their updates to the server. Their updates are thus different from the attackers ones, but not sufficiently similar between themselves in order to be grouped in a single cluster. For the rest of the experiments, loyal clients are generally grouped together because the attacked model is distant from not only the global objective but from all local objectives. Clients have thus more difficulties to reach their local objective, and resemble more between themselves because they begin the new training at the same distant starting point.
 
\section{Conclusion and future work}
In this work, we apply clustering techniques to FL under heterogeneous data. During FedAvg, we exploit the available information sent by the sampled clients at each round to compute similarities between clients incrementally, which enables us to cluster clients without having to compute \textit{all} their parameters at the same round. Our method is especially advantageous in a cross-device context where the number of clients is large, and the communication costs between them and the server are very high. We empirically show on a variety of non-IID settings that the obtained groups reach IID data performances. We also obtain partitions that effectively group clients following similar data distributions if most clients have performed enough rounds of local optimization. 
Moreover, we also address attacks on models as a form of data heterogeneity and apply our method as a defense technique consisting in separating malicious clients from the rest of the clients. We show that our method is a robust defense even when the malicious clients are in the majority, whereas existing methods fail to protect models in this case. 
Ongoing work consists in providing convergence proofs of our method. Other relevant future work is to check the adaptability of our method in differential privacy contexts where noise is added to parameters to reinforce clients' confidentiality \cite{mcmahan2017learning}. 

\bibliographystyle{splncs04}
\bibliography{biblio}

\begin{thebibliography}{10}
\providecommand{\url}[1]{\texttt{#1}}
\providecommand{\urlprefix}{URL }
\providecommand{\doi}[1]{https://doi.org/#1}

\bibitem{blanchard2017machine}
Blanchard, P., El~Mhamdi, E.M., Guerraoui, R., Stainer, J.: Machine learning
  with adversaries: Byzantine tolerant gradient descent. Advances in Neural
  Information Processing Systems  \textbf{30} (2017)

\bibitem{Blondel_2008}
Blondel, V.D., Guillaume, J.L., Lambiotte, R., Lefebvre, E.: Fast unfolding of
  communities in large networks. Journal of Statistical Mechanics: Theory and
  Experiment  \textbf{2008}(10),  P10008 (Oct 2008).
  \doi{10.1088/1742-5468/2008/10/p10008}

\bibitem{Briggs}
Briggs, C., Fan, Z., Andras, P.: Federated learning with hierarchical
  clustering of local updates to improve training on non-iid data. In: 2020
  International Joint Conference on Neural Networks (IJCNN). pp.~1--9. IEEE
  (2020)

\bibitem{chen2021fedhealth}
Chen, Y., Qin, X., Wang, J., Yu, C., Gao, W.: Fedhealth: A federated transfer
  learning framework for wearable healthcare. IEEE Intelligent Systems
  \textbf{35}(4),  83--93 (2020)

\bibitem{Ghosh}
Ghosh, A., Chung, J., Yin, D., Ramchandran, K.: An efficient framework for
  clustered federated learning. Advances in Neural Information Processing
  Systems  \textbf{33},  19586--19597 (2020)

\bibitem{mhamdi2018hidden}
Guerraoui, R., Rouault, S., et~al.: The hidden vulnerability of distributed
  learning in byzantium. In: International Conference on Machine Learning. pp.
  3521--3530. PMLR (2018)

\bibitem{hard2019federated}
Hard, A., Rao, K., Mathews, R., Ramaswamy, S., Beaufays, F., Augenstein, S.,
  Eichner, H., Kiddon, C., Ramage, D.: Federated learning for mobile keyboard
  prediction. arXiv preprint arXiv:1811.03604  (2018)

\bibitem{hsieh2020noniid}
Hsieh, K., Phanishayee, A., Mutlu, O., Gibbons, P.: The non-iid data quagmire
  of decentralized machine learning. In: International Conference on Machine
  Learning. pp. 4387--4398. PMLR (2020)

\bibitem{hu2021challenges}
Hu, S., Lu, J., Wan, W., Zhang, L.Y.: Challenges and approaches for mitigating
  byzantine attacks in federated learning. arXiv preprint arXiv:2112.14468
  (2021)

\bibitem{jiang2019improving}
Jiang, Y., Kone{\v{c}}n{\`y}, J., Rush, K., Kannan, S.: Improving federated
  learning personalization via model agnostic meta learning. arXiv preprint
  arXiv:1909.12488  (2019)

\bibitem{kairouz2021advances}
Kairouz, P., McMahan, H.B., Avent, B., Bellet, A., Bennis, M., Bhagoji, A.N.,
  Bonawitz, K., Charles, Z., Cormode, G., Cummings, R., D'Oliveira, R.G.L.,
  Eichner, H., Rouayheb, S.E., Evans, D., Gardner, J., Garrett, Z., Gascón,
  A., Ghazi, B., Gibbons, P.B., Gruteser, M., Harchaoui, Z., He, C., He, L.,
  Huo, Z., Hutchinson, B., Hsu, J., Jaggi, M., Javidi, T., Joshi, G., Khodak,
  M., Konečný, J., Korolova, A., Koushanfar, F., Koyejo, S., Lepoint, T.,
  Liu, Y., Mittal, P., Mohri, M., Nock, R., Özgür, A., Pagh, R., Raykova, M.,
  Qi, H., Ramage, D., Raskar, R., Song, D., Song, W., Stich, S.U., Sun, Z.,
  Suresh, A.T., Tramèr, F., Vepakomma, P., Wang, J., Xiong, L., Xu, Z., Yang,
  Q., Yu, F.X., Yu, H., Zhao, S.: Advances and open problems in federated
  learning. arXiv preprint arXiv:1912.04977  (2019)

\bibitem{Karimireddy2020}
Karimireddy, S., Kale, S., Mohri, M., Reddi, S., Stich, S., Suresh, A.:
  {SCAFFOLD}: Stochastic controlled averaging for federated learning. In:
  Proceedings of the 37th International Conference on Machine Learning (ICML).
  vol.~119, pp. 5132--5143 (2020)

\bibitem{lecun2010mnist}
LeCun, Y., Bottou, L., Bengio, Y., Haffner, P.: Gradient-based learning applied
  to document recognition. Proceedings of the IEEE  \textbf{86}(11),
  2278--2324 (1998)

\bibitem{Li2018}
Li, T., Sahu, A.K., Zaheer, M., Sanjabi, M., Talwalkar, A., Smith, V.:
  Federated optimization in heterogeneous networks. Proceedings of Machine
  Learning and Systems  \textbf{2},  429--450 (2020)

\bibitem{marfoq21}
Marfoq, O., Neglia, G., Bellet, A., Kameni, L.: Federated multi-task learning
  under a mixture of distributions. Advances in Neural Information Processing
  Systems  (2021)

\bibitem{mcmahan2017communicationefficient}
McMahan, B., Moore, E., Ramage, D., Hampson, S., y~Arcas, B.A.:
  Communication-efficient learning of deep networks from decentralized data.
  In: Artificial intelligence and statistics. pp. 1273--1282. PMLR (2017)

\bibitem{mcmahan2017learning}
McMahan, H.B., Ramage, D., Talwar, K., Zhang, L.: Learning differentially
  private recurrent language models. In: International Conference on Learning
  Representations (2018)

\bibitem{mishchenko22}
Mishchenko, K., Malinovsky, G., Stich, S., Richtárik, P.: Proxskip: Yes! local
  gradient steps provably lead to communication acceleration! finally! arXiv
  preprint arXiv:2202.09357  (2022)

\bibitem{sattler2019clustered}
Sattler, F., M{\"u}ller, K.R., Samek, W.: Clustered federated learning:
  Model-agnostic distributed multitask optimization under privacy constraints.
  IEEE transactions on neural networks and learning systems  \textbf{32}(8),
  3710--3722 (2020)

\bibitem{smith2018federated}
Smith, V., Chiang, C.K., Sanjabi, M., Talwalkar, A.S.: Federated multi-task
  learning. Advances in neural information processing systems  \textbf{30}
  (2017)

\bibitem{yin2021byzantinerobust}
Yin, D., Chen, Y., Kannan, R., Bartlett, P.: Byzantine-robust distributed
  learning: Towards optimal statistical rates. In: International Conference on
  Machine Learning. pp. 5650--5659. PMLR (2018)

\bibitem{yu2020salvaging}
Yu, T., Bagdasaryan, E., Shmatikov, V.: Salvaging federated learning by local
  adaptation. arXiv preprint arXiv:2002.04758  (2020)

\bibitem{zhao2018federated}
Zhao, Y., Li, M., Lai, L., Suda, N., Civin, D., Chandra, V.: Federated learning
  with non-iid data. arXiv preprint arXiv:1806.00582  (2018)

\end{thebibliography}
\end{document}